%% file: ICASSP_2027.tex
\documentclass[11pt]{article}

\usepackage[letterpaper,margin=1in]{geometry}
\usepackage{amsmath,amssymb,amsthm,mathtools}
\usepackage{newtxtext,newtxmath}
\usepackage{bm}
\usepackage{graphicx}
\usepackage{booktabs}
\usepackage{microtype}
\usepackage{subcaption}
\usepackage[hidelinks]{hyperref}
\usepackage{fancyhdr}

\newcommand{\head}[1]{\smallskip\noindent\textbf{#1.}\hspace{0.45em}}

\newenvironment{keywords}{%
  \vspace{.5em}\noindent{\bfseries\textit{Index Terms}---\,\relax}}{\par}

\theoremstyle{plain}
\newtheorem{theorem}{Theorem}
\newtheorem{proposition}{Proposition}
\newtheorem{lemma}{Lemma}
\newtheorem{corollary}{Corollary}
\theoremstyle{definition}

\newtheorem{remark}{Remark}
\newcommand{\R}{\mathbb{R}}
\newcommand{\E}{\mathbb{E}}
\newcommand{\KL}{D_{\mathrm{KL}}}
\newcommand{\TV}{d_{\mathrm{TV}}}
\newcommand{\calA}{\mathcal{A}}
\newcommand{\calC}{\mathcal{C}}
\newcommand{\calF}{\mathcal{F}}
\newcommand{\calI}{\mathcal{I}}
\newcommand{\calT}{\mathcal{T}}
\newcommand{\calV}{\mathcal{V}}
\newcommand{\calZ}{\mathcal{Z}}
\newcommand{\Ch}{\operatorname{Ch}}
\newcommand{\an}{\operatorname{an}}
\newcommand{\Path}{\operatorname{Path}}
\newcommand{\ind}[1]{\mathbf{1}\{#1\}}
\newcommand{\Var}{\operatorname{Var}}
\newcommand{\Cov}{\operatorname{Cov}}
\DeclareMathOperator*{\argmax}{arg\,max}

\title{A Hidden Markov Preference Model for Online\\ Prediction of Adversary Actions%
\thanks{Preprint. Extended version with complete proofs and additional
experimental detail.}}

\author{Angel Daruna \and Michael Cogswell \and Tixiao Shan \and Yi Yao
  \and Han-Pang Chiu \and Ajay Divakaran\\[0.3em]
  SRI International}

\date{}

\begin{document}

\maketitle
\thispagestyle{fancy}

\begin{abstract}
A defender anticipating an attacker needs a probability distribution over the attacker's next action from information available before that action occurs. Two structures complicate this: the attacker's behavioral regime persists and can switch within an episode, and each action combines an operation with state-dependent arguments whose feasible values change over time. We capture both with a hidden Markov model whose latent state is the active regime and whose regime-conditioned action law is defined on a feasibility-masked tree of executable actions. The same likelihood forms the next-action distribution and updates the regime belief once the action is seen, and a partially observed prefix contributes its resolved local factors. Estimation separates into convex logistic fits that Baum--Welch reuses when regimes are unlabeled. We derive a reach-weighted nodewise KL expression for regime information, a sequence-wise log-loss bound for switching regime paths, and a uniform fallback guarantee. On the CAGE~2 cyber defense benchmark the filtered latent state is worth $0.8$ nats per step, and after a controller switch the loss begins to fall in three steps against ten for a recurrent baseline.
\end{abstract}

\begin{keywords}
Hidden Markov models, discrete choice, online prediction, preference learning, cyber defense
\end{keywords}

\section{Introduction}\label{sec:intro}
An observer often needs a probability distribution over another agent's next action from its own observation record, with a network defender anticipating an intruder as the running example. Because the record omits the agent's private state and objectives, prediction must rely on structure reflected in observable behavior. Behavioral patterns persist across consecutive steps and can change within an episode, as when an intruder moves from reconnaissance to exploitation or reacts to a defensive move, and each action combines an operation with arguments whose admissible values depend on the current state, leaving much of the nominal action space infeasible.

Preference learning estimates a probabilistic choice rule from revealed choices through Bradley--Terry and Luce likelihoods~\cite{bradley1952,mcfadden1974}, and quantal-response versions are fitted to observed attacks in security games~\cite{yang2011,nguyen2013}. Hidden context can cause a single fitted choice law to aggregate behavior generated under different unobserved conditions~\cite{siththaranjan2024}. Latent-class and mixture models capture such heterogeneity at the user or episode level~\cite{train2009,poddar2024}, while input--output hidden Markov models~\cite{bengio1995iohmm} and hidden-Markov choice models~\cite{netzer2008} let the latent class evolve over time. Opponent-modeling methods infer opponent types, goals, or latent representations from interaction~\cite{albrecht2018,nashed2022,rosman2016,papoudakis2021}, and cyber-forecasting methods predict attack stages or red actions from streaming observations~\cite{holgado2020,nadeem2022,li2023,samaddar2026}. The present setting instead requires a full next-action distribution committed before each outcome, with within-episode regime changes, state-dependent feasibility, and evaluation under sequential log loss.

We represent the active behavioral regime with a conditional hidden Markov model~\cite{bengio1995iohmm,rabiner1989} and structured executable actions with a regime-conditioned choice law on the feasible action tree. The same likelihood forms the next-action prediction before the action occurs and updates the regime belief afterwards; for a tree-prefix observation, the resolved local factors give its likelihood and the evidence it contributes about the regime. The construction retains exact filtering, and estimation separates into convex fits that Baum--Welch reuses when regimes are unlabeled.

\textbf{Contributions.}
The latent dynamics follow the input--output HMM~\cite{bengio1995iohmm} and hidden-Markov choice models~\cite{netzer2008}, and the log-loss bounds are mixture inequalities of the expert-tracking framework~\cite{merhav1998,herbster1998,bousquet2002,koolen2013}; the new elements are (i)~a regime-conditioned choice law on a feasibility-masked action tree, with shared and regime-specific terms, whose identification rate is a reach-weighted sum of nodewise KL divergences over the observed action fields; (ii)~exact filtering that treats a complete action and a tree-prefix observation through the same local factors; (iii)~switching and uniform-fallback guarantees on the feasible tree; and (iv)~an attacker-side CAGE Challenge~2 evaluation~\cite{kiely2023} of a two-stage form of the predictor that separates action structure, latent-state tracking, and soft regime weighting.


\section{Problem Formulation}\label{sec:problem}

\head{Prediction target}
An \emph{agent} and an \emph{observer} interact at discrete times $t=1,2,\dots$, and the observer predicts each action the agent takes. Let $h_t^{-}$ be the observer history available immediately before the agent commits to $A_t$, including any observer action taken first at step $t$, and $h_t^{+}$ the same history after $A_t$ is revealed. The prediction target is
\begin{equation}
p_t^\star(a)=\Pr(A_t=a\mid h_t^{-}), \qquad x_t=\psi(h_t^{-})\in\R^d,
  \label{eq:target}
\end{equation}
with $\psi$ a fixed feature map to a context vector. Given training episodes, the problem is to produce at each $t$ a distribution $\widehat p_t$ over the feasible actions, committed before the step's outcome is seen, that makes the sequential logarithmic loss $\sum_t-\log\widehat p_t(a_t)$ small on the executed actions. In the \emph{replay} setting logs reveal the complete executed action; in the \emph{detection} setting the observer receives a coarsened observation of it.

\head{Actions and feasibility}
Actions are the leaves of a rooted tree $\calT$, the action grammar, with internal nodes $\calI$, children $\Ch(v)$ at an internal node $v$, and leaf set $\calA$. Each internal node resolves one semantic field of the action, with the operation family at the root and its arguments below. For a leaf $a$, write $j_v(a)$ for the child taken at $v$ on the path to $a$, let $\calI(a)\subseteq\calI$ be the internal nodes on the root-to-$a$ path, and write $\Path(a)=\{(v,j_v(a)):v\in\calI(a)\}$. Feasibility at time $t$ is a mask $m_{t,v}(j)\in\{0,1\}$; the feasible children at $v$ are $\calC_t(v)=\{j\in\Ch(v):m_{t,v}(j)=1\}$, and the feasible leaves $\calA_t$ are those whose entire path is feasible. Masks are pruned bottom-up, so every feasible child leads to at least one feasible leaf, the local choice laws normalize over $\calC_t(v)$, and their product induces a normalized distribution over $\calA_t$. Masks are computed from the grammar and $h_t^{-}$ as conservative supersets of the true support, so every executed action remains feasible under the observer's model and constraints depending on the agent's private progress remain part of the modeled uncertainty.

\head{Latent regime}
The agent's regime at time $t$ is a latent state $Z_t\in\calZ$ with $|\calZ|=K$. The regime is behavioral, for instance a policy family, a stage of an attack, or an operating constraint; it persists over consecutive steps, may change within an episode, and is defined only through its effect on the action distribution. The regime evolves through a context-dependent Markov kernel with an explicit self-transition term,
\begin{equation}
P_{\eta}(z'\mid z,x_t)
=
\frac{
\exp\bigl(c\,\ind{z'=z}+\eta_{z'}^{\top}\bar x_t\bigr)
}{
\sum_{r}
\exp\bigl(c\,\ind{r=z}+\eta_{r}^{\top}\bar x_t\bigr)
},
  \label{eq:transition}
\end{equation}
in which $z'$ indexes $Z_t$, $z$ indexes $Z_{t-1}$, $\bar x_t=[1,x_t^{\top}]^{\top}$, and $c\ge0$ controls persistence. The shared persistence term keeps the transition model small when switch data are sparse, and fixing one $\eta_{z'}$ at zero identifies $(c,\eta)$. The transition into $Z_t$ uses $x_t$, which is available before $A_t$ and can depend on evidence accumulated through step $t-1$ together with any observer action taken first at step $t$.

Setting $\eta_{z'}\equiv0$ gives the homogeneous \emph{persistent} kernel
\begin{equation}
P(z'\mid z)
=
\kappa\,\ind{z'=z}
+
\tfrac{1-\kappa}{K-1}\,\ind{z'\neq z},
\qquad
\kappa\in[\tfrac1K,1),
  \label{eq:sticky}
\end{equation}
with $\kappa=e^{c}/(e^{c}+K-1)$ and expected dwell time $1/(1-\kappa)$. The pathwise log-loss result applies to the general kernel in \eqref{eq:transition}, while the explicit switching, fallback, and adaptation bounds use \eqref{eq:sticky}. At $c=0$ every row of \eqref{eq:transition} equals the contextual gate $\mathrm{softmax}_{z'}(\eta_{z'}^{\top}\bar x_t)$, so the filter carries no information across steps and the model reduces to a memoryless mixture of experts with a contextual gate.

With $\eta=(c,\{\eta_{z'}\})$ and $\Theta$ the action-model parameters, \eqref{eq:transition} and the regime-conditioned action law of Section~\ref{sec:model} define an HMM~\cite{bengio1995iohmm,rabiner1989} whose joint probability, for a predictable context sequence $x_{1:T}$, regime path $z_{1:T}$, and action sequence $a_{1:T}$, is
\begin{equation}
\nu(z_1)\,p_{\Theta}(a_1\mid x_1,z_1)
\prod_{t=2}^{T}
P_{\eta}(z_t\mid z_{t-1},x_t)\,
p_{\Theta}(a_t\mid x_t,z_t),
  \label{eq:joint}
\end{equation}
with initial distribution $\nu$.

\section{Model, Inference, and Learning}\label{sec:model}

\head{Action model}
Conditional on $Z_t=z$ and context $x$, the action follows a choice model assembled from local decisions at the internal nodes. Each child $j\in\Ch(v)$ carries a descriptor $e_{v,j}\in\R^{r_v}$, and we combine context and child identity through $\phi_v(x,e_{v,j})=\bar x\otimes e_{v,j}$, allowing the same context to change the relative preference among children at a node. The local score is
\begin{equation}
  u_{v,z,j}(x)=\big\langle \theta_v^{0}+\Delta_{v,z},\,
  \phi_v(x,e_{v,j})\big\rangle,
  \label{eq:score}
\end{equation}
with the vector $\theta_v^{0}$ shared across regimes and $\Delta_{v,z}$ the per-regime residual. The local choice and the action probabilities are
\begin{align}
  \pi_{v,z}(j\mid x)&=
    \frac{\exp u_{v,z,j}(x)}
    {\sum_{k\in\calC_t(v)}\exp u_{v,z,k}(x)},
  \label{eq:local}\\
  p_{\Theta}(a\mid x,z)&=\!\!\prod_{(v,j_v(a))\in\Path(a)}\!\!
    \pi_{v,z}(j_v(a)\mid x),\quad a\in\calA_t,
  \label{eq:leaf}
\end{align}
and zero for $a\notin\calA_t$, with the time dependence induced by the mask left implicit. The factorization \eqref{eq:leaf} is a hierarchical softmax~\cite{morin2005} with locally normalized choices along the feasible action path; it yields separate supervised likelihood terms at each node and decomposes regime differences across the visited nodes. The decomposition $\theta_v^{0}+\Delta_{v,z}$ implements partial pooling across regimes that reach the same node: the shared term carries the common choice structure and $\Delta_{v,z}$ the regime-specific departures, which are the only source of evidence about the active regime, so shrinking the residuals increases pooling while reducing identification; Section~\ref{sec:theory} quantifies this trade-off. We impose $\sum_z\Delta_{v,z}=0$ and use centered descriptors and a reference child, which identify $\theta_v^{0}$ as the across-regime mean and, with the quadratic penalties, make each supervised node problem strictly convex.

\head{Prediction}
Let $b_t^{-}(z)=\Pr(Z_t=z\mid h_t^{-})$ be the posterior over the hidden regime before the agent acts and $b_t^{+}(z)=\Pr(Z_t=z\mid h_t^{+})$ the posterior after. The predictor is the one-step predictive distribution of the hidden Markov model, obtained by averaging the regime-conditioned action laws over the current belief,
\begin{equation}
  \widehat p_t(a)=\sum_{z\in\calZ}b_t^{-}(z)\,p_{\Theta}(a\mid x_t,z),
  \qquad a\in\calA_t,
  \label{eq:pred}
\end{equation}
and the belief obeys, with $b_1^{-}=\nu$ and evidence weight $\ell_t(z)$ defined below,
\begin{align}
  b_t^{+}(z)&=
  \frac{\ell_t(z)\,b_t^{-}(z)}
       {\sum_{r}\ell_t(r)\,b_t^{-}(r)},
  \label{eq:update}\\
  b_{t+1}^{-}(z')&=
  \sum_{z}P_{\eta}(z'\mid z,x_{t+1})\,b_t^{+}(z).
  \label{eq:time}
\end{align}
In replay $\ell_t(z)=p_{\Theta}(a_t\mid x_t,z)$, and in detection $\ell_t(z)$ is the likelihood $\sum_{a\in\calA_t}\omega(e_t\mid a)\,p_{\Theta}(a\mid x_t,z)$ induced by the observation channel $\omega$, with $\ell_t\equiv1$ on an evidence-free step; for a tree-prefix observation it reduces to the product of the resolved local choice probabilities. The prediction \eqref{eq:pred} is committed before $\ell_t$ is formed and therefore uses only $h_t^{-}$. One step runs \eqref{eq:time}, one feasible-tree pass per regime for \eqref{eq:local}--\eqref{eq:leaf}, the prediction \eqref{eq:pred}, and the update \eqref{eq:update}, in time linear in $K$ under \eqref{eq:sticky} and in the size of the feasible tree; Appendix~\ref{app:verify} gives the per-step cost in full and verifies that $b_t^{+}$ is a sufficient statistic of the history.

\head{Estimation}
When regime labels are observed, the penalized negative log-likelihood of one episode is
\begin{equation}
\begin{aligned}
  -\log\nu(z_1)
  &-\!\sum_{t=2}^{T}\!\log P_{\eta}(z_t\mid z_{t-1},x_t)\\[-1pt]
  &-\!\sum_{t=1}^{T}\!\sum_{(v,j)\in\Path(a_t)}\!\!\!\log\pi_{v,z_t}(j\mid x_t),
\end{aligned}
  \label{eq:supervised}
\end{equation}
summed over episodes with quadratic regularizers on $(c,\eta)$ and $\{\theta_v^{0},\Delta_{v,z}\}$. Every term is the negative log of a softmax linear in the parameters, so with observed regimes the objective is jointly convex and separates into one transition fit and one choice fit per node, each solved by L-BFGS-B under $c\ge0$. With latent regimes the marginal likelihood is nonconvex, and Baum--Welch alternates exact forward--backward responsibilities with weighted versions of the same convex subproblems, increasing the penalized marginal likelihood monotonically to a stationary point. Held-out predictive likelihood selects $K$. Single-regime trajectories contain no observed switches, so persistence is selected by validation NLL on separate state-preserving switch episodes. Appendix~\ref{app:verify} records the convexity, separability, and uniqueness argument for \eqref{eq:supervised} and the role of the residual in identifiability.

\head{Uniform fallback state}
For a finite reference loss that does not depend on the fit of the learned regimes, we add a state $\bot$ with node probabilities $\pi_{v,\bot}(j\mid x)=N_t(j)/N_t(v)$, where $N_t(v)$ counts the feasible leaves below $v$. These telescope to $p_{\Theta}(a\mid x,\bot)=1/|\calA_t|$, uniform over the feasible actions, while preserving the product form \eqref{eq:leaf}. As an ordinary state of \eqref{eq:sticky} on $K{+}1$ states, $\bot$ gives the fallback guarantee of Section~\ref{sec:theory}, and $b_t^{+}(\bot)$ is the weight assigned to it.

\section{Analysis}\label{sec:theory}
Conditioning on the fitted parameters, we first quantify how the resolution of an observed action determines the evidence it provides about the active regime, then compare cumulative predictive loss with regime paths that may switch. The arguments are short: \eqref{eq:coarse} sums the local laws below $v$ from the leaves up, \eqref{eq:ident} is the chain rule for relative entropy on the tree, and \eqref{eq:track} keeps one path of the forward recursion. Complete proofs are given in Appendices~\ref{app:prefix}--\ref{app:cor2}, under the conventions of Appendix~\ref{app:conv}.

\head{Information from partial observations}
\begin{lemma}[Tree-prefix likelihood]\label{lem:prefix}
For a node $v$ with ancestors $\an(v)$, writing
$\calA_t(v)=\{a\in\calA_t:v\in\calI(a)\}$,
\begin{equation}
\sum_{a\in\calA_t(v)}p_{\Theta}(a\mid x,z)
=
\prod_{v'\in\an(v)}
\pi_{v',z}\bigl(j_{v'}(v)\mid x\bigr),
  \label{eq:coarse}
\end{equation}
where $j_{v'}(v)$ is the child of $v'$ toward $v$. Taking $v$ as the root gives normalization over $\calA_t$.
\end{lemma}
An observation that resolves the action path to depth $k$ therefore has likelihood equal to the product of the $k$ resolved local factors, and each resolved node can contribute evidence about the active regime (proof in Appendix~\ref{app:prefix}).

\begin{corollary}[Regime identification rate]\label{cor:ident}
Fix $x$ and the feasible tree, let the executed action be drawn from regime $z^{\star}$, and let the evidence resolve the executed path to depth $k$, with $\calV_k$ the set of feasible internal nodes at depth less than $k$. For every regime $z$, the expected log-likelihood ratio entering \eqref{eq:update} is
\begin{equation}
\E\Bigl[\log\tfrac{\ell_t(z^{\star})}{\ell_t(z)}\Bigr]
=
\sum_{v\in\calV_k}
\rho_{v,z^{\star}}\,
\KL\bigl(\pi_{v,z^{\star}}\,\|\,\pi_{v,z}\bigr),
  \label{eq:ident}
\end{equation}
with $\rho_{v,z^{\star}}$ the probability under $z^{\star}$ that the path reaches $v$. At full resolution the sum equals
$\KL(p_{\Theta}(\cdot\mid x,z^{\star})\,\|\,p_{\Theta}(\cdot\mid x,z))$,
so unresolved nodes remove exactly their terms from the expected one-step increase of the posterior log-odds.
\end{corollary}
Each local term is, to second order, a quadratic form in the difference between regime residuals, so residual shrinkage lowers the identification rate, quantifying the trade-off between pooling and regime separation; Appendix~\ref{app:ident} proves the corollary and Appendix~\ref{app:verify} derives the second-order form.

\head{Log-loss guarantees}
\begin{theorem}[Individual-sequence tracking]\label{thm:track}
Let $\widehat p_t$ be the predictor \eqref{eq:pred} produced by \eqref{eq:update}--\eqref{eq:time} with fixed parameters on the replay sequence. For every predictable context sequence $x_{1:T}$, every realized action sequence satisfying $a_t\in\calA_t$, and every regime path $z_{1:T}\in\calZ^{T}$,
\begin{equation}
\begin{aligned}
  \sum_{t=1}^{T}-\log\widehat p_t(a_t)
  \le
  \sum_{t=1}^{T}&-\log p_{\Theta}(a_t\mid x_t,z_t)\\[-1pt]
  &-\log\nu(z_1)
  -\!\sum_{t=2}^{T}\!\log P_{\eta}(z_t\mid z_{t-1},x_t).
\end{aligned}
  \label{eq:track}
\end{equation}
\end{theorem}
The detection setting obeys the same inequality with $\ell_t$ in place of the action likelihood. Minimizing the right-hand side over $z_{1:T}$ compares the deployed predictor with the regime path of smallest action loss plus its initial and transition cost under the fitted kernel. Appendix~\ref{app:track} gives the proof and the corresponding detection-setting inequality.

\begin{corollary}[Fixed share and uniform fallback]\label{cor:safe}
With uniform $\nu$ and the homogeneous kernel \eqref{eq:sticky}, a comparator path with $S$ switches gives
\begin{equation}
\begin{aligned}
  \sum_{t=1}^{T}-\log\widehat p_t(a_t)
  \le
  \sum_{t=1}^{T}&-\log p_{\Theta}(a_t\mid x_t,z_t)+\log K\\[-1pt]
  &+(T{-}1{-}S)\log\tfrac1\kappa
  +S\log\tfrac{K-1}{1-\kappa},
\end{aligned}
  \label{eq:sticky_bound}
\end{equation}
the fixed-share penalty for tracking changing experts~\cite{herbster1998,cesabianchi2006} with the regime-conditioned action laws as the experts. If $\bot$ is an ordinary state of \eqref{eq:sticky} on $K{+}1$ regimes, the constant comparator $z_t\equiv\bot$ gives, for every feasible action sequence,
\begin{equation}
  \sum_{t=1}^{T}-\log\widehat p_t(a_t)
  \le
  \sum_{t=1}^{T}\log|\calA_t|
  +\log(K{+}1)
  +(T{-}1)\log\tfrac1\kappa.
  \label{eq:safe}
\end{equation}
\end{corollary}
A single $\kappa$ controls both penalties: by \eqref{eq:safe} the mean loss exceeds uniform coding over the feasible sets by at most $\log\tfrac1\kappa+\log(K{+}1)/T$, so no feasible action sequence is catastrophic, while carrying $\bot$ costs a learned-regime comparator only $\log\tfrac{K+1}{K}+S\log\tfrac{K}{K-1}$ per episode; Appendix~\ref{app:cor2} proves both bounds and Appendix~\ref{app:verify} works the trade-off out numerically. Greater persistence lowers the penalty for staying in one regime but raises the prior evidence to be overcome after a switch; since \eqref{eq:sticky} floors every $b_t^{-}(z')$ at $\tfrac{1-\kappa}{K-1}$, a Wald-type argument bounds the expected time for the posterior log-odds to favor the new regime by $\bigl(\log\tfrac{K-1}{1-\kappa}+B\bigr)/r$, with $r$ a floor on the evidence rate \eqref{eq:ident} and $B$ a bound on a single-step log-likelihood ratio, stated and proved as Remark~\ref{rem:delay} in Appendix~\ref{app:delay}.

\begin{figure}[!t]
\centering
\begin{subfigure}[t]{0.48\textwidth}
\centering
\includegraphics[width=\linewidth]{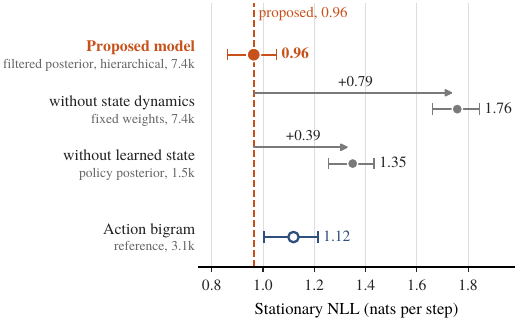}
\caption{}
\label{subfig:ablation}
\end{subfigure}
\hfill
\begin{subfigure}[t]{0.48\textwidth}
\centering
\includegraphics[width=\linewidth]{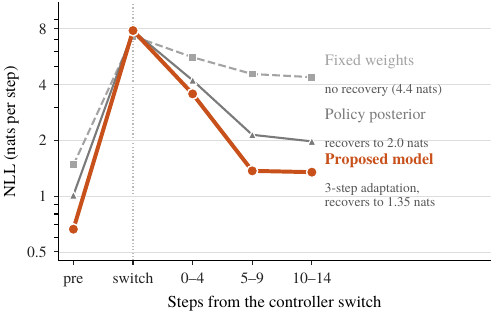}
\caption{}
\label{subfig:switch}
\end{subfigure}
\caption{(a) Stationary test set, blue-observation context, $90\%$ intervals; each ablation removes one element of the regime posterior and the arrow gives the change in NLL. (b) Mean NLL before a controller switch, at the switch step, and in three five-step windows after it, on a log scale.}
\label{fig:results}
\end{figure}

\section{Experiments}\label{sec:exp}

\head{Setup}
We evaluate on CAGE Challenge~2~\cite{kiely2023,cage2} in the bug-fixed CybORG++ build~\cite{emerson2024,cyborg2021}, where a red agent attacks a simulated enterprise network against a fixed reactive blue baseline, a setting studied so far mainly from the defender's side~\cite{foley2022,hammar2024,vyas2025}; the closest precedent, \cite{samaddar2026}, imitates a red policy from network observations and reports accuracy. The action tree has the operation at the root (no-op, system discovery, service discovery, exploitation, escalation, impact) and one operation-dependent argument below it, a subnet for system discovery and a host otherwise, giving $56$ executable actions. The generating red policies B\_line, Meander, and Sleep are the policy classes; Sleep is a point mass on no-op, so all predictors are scored on active steps, those whose executed action is not no-op. We train on $40$ episodes of length $100$ per policy, validate on $10$, and hold out $25$ per policy as the stationary test set. Switch episodes replace the red controller once at $\tau\sim\mathrm{Unif}\{30,\dots,70\}$ with session state preserved; $60$ of them select the persistence $\kappa$ of the policy-class chain by validation NLL and a disjoint set of $N$ scores regime tracking. The context $x_t$ holds the observer-available coordinates of the blue observation and features computed from them before the current red action, and the full context adds the previous red action as a one-hot vector. All runs use the replay setting. Metrics are per-step NLL in nats over the feasible set, top-$1$ accuracy, and $10$-bin ECE of the top-$1$ confidence, with $90\%$ episode-level bootstrap intervals.

\head{Estimator and comparisons}
The evaluated estimator is a two-stage form of the predictor with convex fits and inputs shared by every compared choice model. A $48$-state HMM over actions with an unconstrained transition matrix is fit to the training action sequences without labels, and its filtered posterior $b_t^{-}$ enters the node scores of the masked choice model linearly beside the context, one convex logistic regression per node. The posterior thus pools the regime laws in the scores, which coincides with \eqref{eq:pred} at a one-hot posterior; the choice likelihood does not update the chain, so the bounds of Section~\ref{sec:theory} describe the joint filter \eqref{eq:pred} and are not claimed for this estimator. Ablations replace the filtered posterior by fixed state weights, by a one-hot encoding of its mode, or by nothing; the hierarchical choice model by a flat softmax over the $56$ actions with the same inputs; and the learned chain by a three-state chain whose states are the policy classes with kernel \eqref{eq:sticky}. Reference predictors are a first-order Markov model on actions (the action bigram) and a masked GRU seeing the full context and mask and carrying its own recurrent state.

\begin{table}[t]
\centering
\caption{Stationary test set ($25$ held-out episodes per policy, active steps) and switch test set. NLL in nats per step, top-$1$ in percent, $10$-bin ECE; stationary NLL carries its $90\%$ bootstrap interval and switch intervals are about $\pm0.3$ nats. Choice models are hierarchical unless marked flat. The second group uses the blue-observation context, the third adds the previous action, and bold rows are the proposed model.}
\label{tab:main}
\small
\setlength{\tabcolsep}{4pt}
\renewcommand{\arraystretch}{0.95}
\begin{tabular}{lccccc}
\toprule
 & \multicolumn{3}{c}{Stationary} & \multicolumn{2}{c}{Switch}\\
\cmidrule(lr){2-4}\cmidrule(lr){5-6}
Predictor & NLL$\downarrow$ [CI] & Top-1$\uparrow$ & ECE$\downarrow$ & NLL$\downarrow$ & ECE$\downarrow$\\
\midrule
Action bigram               & 1.119 [1.00, 1.22] & 63.8 & 0.037 & 1.946 & 0.077\\
Masked GRU                  & 0.872 [0.76, 0.96] & 70.5 & 0.054 & 1.821 & 0.112\\
\midrule
Fixed state weights         & 1.758 [1.66, 1.85] & 44.7 & 0.186 & 2.778 & 0.232\\
Policy-class posterior      & 1.350 [1.26, 1.43] & 56.7 & 0.074 & 2.634 & 0.125\\
Posterior mode              & 0.958 [0.85, 1.05] & 67.4 & 0.052 & 2.142 & 0.107\\
Filtered posterior, flat    & 0.917 [0.81, 1.01] & 68.7 & 0.059 & 1.899 & 0.105\\
\textbf{Filtered posterior} & \textbf{0.964} [0.86, 1.05] & 67.0 & 0.047 & 1.971 & 0.088\\
\midrule
No latent state             & 1.068 [0.97, 1.16] & 63.9 & 0.052 & 2.122 & 0.089\\
Policy-class posterior      & 1.033 [0.93, 1.12] & 64.5 & 0.042 & 2.201 & 0.088\\
\textbf{Filtered posterior} & \textbf{0.924} [0.82, 1.01] & 68.0 & 0.045 & 1.951 & 0.084\\
\bottomrule
\end{tabular}
\end{table}

\head{Stationary prediction}
Table~\ref{tab:main} and Figure~\ref{fig:results}(a) separate the two design choices. Removing the state dynamics costs $0.79$ nats per step and $22$ points of top-$1$ accuracy, and replacing the learned chain by the three policy classes costs $0.39$, or $0.11$ with the full context. The learned states thus carry information beyond the policy label, part of which the previous action supplies. Replacing the posterior by its mode changes nothing while the regime is stable. The flat softmax gains $0.047$ nats at $44\%$ more parameters and is worse calibrated. The GRU, at $7{,}288$ parameters, reaches $0.872$ nats and $70.5\%$ against $0.924$ and $68.0\%$ for the full-context model, with overlapping intervals, and the choice model is the better calibrated of the two. The action bigram is a strong reference at $1.119$ nats because B\_line is close to deterministic given its last action, while uniform coding of the $56$ actions costs $4.03$. Over a sweep of the training-set size, the hierarchical model is at or below the flat one up to $60$ episodes and behind it from $90$ on. Appendix~\ref{app:verify} lists the parameter counts of every predictor and the training-set sweep in full.

\head{Regime tracking}
The switch columns of Table~\ref{tab:main} and Figure~\ref{fig:results}(b) report the switch episodes. At the switch step every predictor loses between $7.3$ and $9.5$ nats; the filtered hierarchical model loses $7.81$, against $8.13$ for its mode and $9.23$ for the GRU, and its loss begins to fall after three steps against ten for the GRU. Over the whole episode, weighting by the full posterior is worth $0.17$ nats per step over its mode, the same margin the full-context model holds over no latent state, with overlapping intervals in both cases, though the paired per-episode differences exclude zero. Ten to fourteen steps after the switch, fixed weights stay at $4.4$ nats and the policy-class chain at $2.0$, against $1.35$ for the proposed model; the GRU recovers further, to $1.05$ in that window, and ends the episode at $1.821$ against $1.951$, again inside the intervals. Calibration favors the proposed model under switching as well, ECE $0.084$ against $0.112$ for the GRU and $0.105$ for the flat softmax.

\head{Summary}
Across both experiments the accuracy of the next-action distribution is set by how well the predictor represents \emph{where the attacker is} rather than by the capacity of the action model: the filtered latent state is worth $0.8$ nats per step over fixed weights and $0.4$ over the known policy classes, whereas the hierarchical and flat choice models are separated by $0.047$ and the same-size GRU by less than one interval width. What the latent state buys is a calibrated belief rather than a label, and this is where the structure pays: weighting by the full posterior matches its mode in steady state but gains $0.17$ nats per step under switching, and the loss begins to fall three steps after a controller change against ten for the GRU, at $30\%$ fewer parameters than the flat model and better calibration than both baselines. The two open directions follow from the two places the belief is still weakest. Fitting the chain and the choice law jointly, rather than in the two stages evaluated here, would let the action likelihood sharpen the regime posterior and would bring the guarantees of Section~\ref{sec:theory} to bear on the deployed predictor, including the fallback state $\bot$ whose absence explains the overconfidence observed against a held-out red policy (Corollary~\ref{cor:safe}, with the measurements in Appendix~\ref{app:verify}). Beyond replay, the same local factors score a partially resolved action, so \eqref{eq:ident} predicts how much regime evidence survives coarse detections; testing that in the detection setting is the natural next step towards a defender that acts on the prediction.

\section*{Acknowledgments}

This research was developed with funding from the Defense Advanced Research Projects Agency (DARPA). The views, opinions and/or findings expressed are those of the authors and should not be interpreted as representing the official views or policies of DARPA or the U.S. Government.

\bibliographystyle{IEEEbib}
\bibliography{REF}

\clearpage
\appendix
\numberwithin{equation}{section}

\phantomsection
\pdfbookmark[0]{Appendices}{appendices}
\begin{center}
  \rule{0.85\textwidth}{0.4pt}\\[0.9em]
  {\LARGE\bfseries Appendices}\\[0.8em]
  {\large Proofs, Omitted Results, and Additional Experimental Detail}\\[0.9em]
  \rule{0.85\textwidth}{0.4pt}
\end{center}
\vspace{1.5em}

\input{ICASSP_appendix}

\end{document}

%% file: ICASSP_appendix.tex

\noindent These appendices collect complete proofs of the results stated in the
body, state the results that were omitted there for length, verify the
identities asserted in the text, and record the additional experimental detail
referenced from it. Equations numbered $(1)$--$(16)$ are those of the body;
equations numbered by appendix section, such as \eqref{eq:reach}, are local to
the appendices.

\section{Conventions Used in the Proofs}\label{app:conv}
All parameters $(\nu,\eta,\Theta)$ are fixed, and $t$, the context $x$, and the masks are fixed wherever a single step is analyzed. Masks are pruned bottom-up, as stated in Section~\ref{sec:problem}: every feasible child of a node on a feasible path retains at least one feasible leaf in its subtree, so the sets $\calC_t(v)$ are the child sets of a pruned feasible tree whose leaf set is $\calA_t$. Every local law is extended by $\pi_{v,z}(j\mid x)=0$ for $j\notin\calC_t(v)$. The softmax \eqref{eq:local} is strictly positive on $\calC_t(v)$, and the background weights $N_t(j)/N_t(v)$ are positive under pruning, so $p_{\Theta}(a\mid x,z)>0$ for every $a\in\calA_t$ and every regime. The kernels \eqref{eq:transition} and \eqref{eq:sticky} with $\kappa<1$ are likewise strictly positive, so every divergence below is finite, with the convention $0\log0=0$. Two classical facts are used. Pinsker's inequality states $\KL(p\,\|\,q)\ge2\,\TV^{2}(p,q)$~\cite{cesabianchi2006}. The log-sum inequality states that for nonnegative reals $a_1,\dots,a_n$ and $b_1,\dots,b_n$ with $A=\sum_i a_i$ and $B=\sum_i b_i>0$, $\sum_i a_i\log(a_i/b_i)\ge A\log(A/B)$; it follows from Jensen's inequality for the convex function $\varphi(t)=t\log t$, since $\sum_i a_i\log(a_i/b_i)=B\sum_i(b_i/B)\,\varphi(a_i/b_i)\ge B\,\varphi(A/B)=A\log(A/B)$.

\section[Proof of Lemma 1]{Proof of Lemma~\ref{lem:prefix}}\label{app:prefix}
\noindent\textbf{Lemma~\ref{lem:prefix} (Tree-prefix likelihood, restated).} \emph{For a node $v$ with ancestors $\an(v)$, writing $\calA_t(v)=\{a\in\calA_t: v \text{ lies on the root-to-}a\text{ path}\}$,}
\begin{equation*}\tag{\ref*{eq:coarse}}
  \sum_{a\in\calA_t(v)}p_{\Theta}(a\mid x,z)
  =\prod_{v'\in\an(v)}\pi_{v',z}\bigl(j_{v'}(v)\mid x\bigr),
\end{equation*}
\emph{where $j_{v'}(v)$ is the child of $v'$ toward $v$, extending the leaf notation $j_v(a)$ to internal targets. Taking $v$ as the root gives normalization over $\calA_t$.}

\begin{proof}
If the path from the root to $v$ leaves the pruned feasible tree, then $\calA_t(v)=\varnothing$ and at least one factor on the right vanishes by the zero-extension convention, so both sides are zero. Assume therefore that $v$ lies on the pruned tree. For a node $u$ of the pruned tree define $R(u):=1$ when $u$ is a leaf and, when $u$ is internal,
\[
  R(u):=\sum_{a\in\calA_t(u)}\
  \prod_{\substack{w\in\calI(a),\ w\ \text{in the}\\ \text{subtree rooted at}\ u}}
  \pi_{w,z}\bigl(j_w(a)\mid x\bigr),
\]
the mass of the subtree below $u$ under the local factors at $u$ and its descendants. We show $R(u)=1$ by induction on the height of $u$ in the pruned tree. If every child in $\calC_t(u)$ is a leaf, then $\calA_t(u)$ consists of the leaf children of $u$, the product reduces to the single factor at $u$, and $R(u)=\sum_{j\in\calC_t(u)}\pi_{u,z}(j\mid x)=1$ by \eqref{eq:local}. For a general internal $u$, the leaves of $\calA_t(u)$ partition according to the child taken at $u$; write $c_j$ for the child of $u$ with index $j$. For a leaf through $c_j$ the product splits as $\pi_{u,z}(j\mid x)$ times the corresponding product below $c_j$, which is empty, hence one, when $c_j$ is a leaf. Pruning guarantees that every child in $\calC_t(u)$ supports at least one feasible leaf, so the inductive hypothesis applies to each child and
\[
  R(u)=\sum_{j\in\calC_t(u)}\pi_{u,z}(j\mid x)\,R(c_j)
      =\sum_{j\in\calC_t(u)}\pi_{u,z}(j\mid x)=1 .
\]
For the lemma, every $a\in\calA_t(v)$ traverses the same ancestor edges $(v',j_{v'}(v))$ with $v'\in\an(v)$, so \eqref{eq:leaf} factors its probability into the fixed ancestor product times the product at and below $v$; summing over $\calA_t(v)$ and applying $R(v)=1$ gives \eqref{eq:coarse}. At the root the ancestor product is empty and \eqref{eq:coarse} states $\sum_{a\in\calA_t}p_{\Theta}(a\mid x,z)=1$.
\end{proof}
The same bookkeeping yields an identity used repeatedly below. For an internal node $v$ of the pruned tree and $j\in\calC_t(v)$, restricting the sum to the leaves that take child $j$ at $v$ isolates the factor $\pi_{v,z}(j\mid x)$ in front of the mass of the subtree below that child, which is one, so with $\rho_{v,z}(x):=\prod_{v'\in\an(v)}\pi_{v',z}(j_{v'}(v)\mid x)$,
\begin{equation}
  \sum_{\substack{a\in\calA_t:\ v\in\calI(a)\\ j_v(a)=j}}
  p_{\Theta}(a\mid x,z)
  =\rho_{v,z}(x)\,\pi_{v,z}(j\mid x).
  \label{eq:reach}
\end{equation}
By \eqref{eq:coarse}, $\rho_{v,z}(x)$ is the probability that the action path reaches $v$; dividing \eqref{eq:reach} by it shows that, conditionally on reaching $v$, the child taken at $v$ has law $\pi_{v,z}(\cdot\mid x)$.

\section[Proof of Corollary 1]{Proof of Corollary~\ref{cor:ident}}\label{app:ident}
\noindent\textbf{Corollary~\ref{cor:ident} (Regime identification rate, restated).}
\emph{Fix $x$ and the feasible tree, let the executed action be drawn from regime $z^{\star}$, and let the evidence resolve the executed path to depth $k$, with $\calV_k$ the set of feasible internal nodes at depth less than $k$. For every regime $z$, the expected log-likelihood ratio entering \eqref{eq:update} is}
\begin{equation*}\tag{\ref*{eq:ident}}
  \E\Bigl[\log\tfrac{\ell_t(z^{\star})}{\ell_t(z)}\Bigr]
  =\sum_{v\in\calV_k}\rho_{v,z^{\star}}\,
   \KL\bigl(\pi_{v,z^{\star}}\,\|\,\pi_{v,z}\bigr),
\end{equation*}
\emph{with all local laws at the context $x$ and $\rho_{v,z^{\star}}$ the probability under $z^{\star}$ that the action path reaches $v$. At full resolution the sum equals $\KL(p_{\Theta}(\cdot\mid x,z^{\star})\,\|\,p_{\Theta}(\cdot\mid x,z))$, so unresolved nodes remove exactly their terms from the expected one-step increase of the posterior log-odds.}
\begin{proof}
Let $A\sim p_{\Theta}(\cdot\mid x,z^{\star})$ and write $J_v:=j_v(A)$ for the child taken at an internal node $v\in\calI(A)$. Prefix evidence to depth $k$ reveals the resolved edges of the executed path: the compatible leaves form $\calA_t(v_k)$ with $v_k$ the deepest resolved node, and when the executed leaf lies at depth at most $k$ the evidence reveals the leaf itself. In either case Lemma~\ref{lem:prefix} gives $\ell_t(z)=\prod_{v\in\calI(A)\cap\calV_k}\pi_{v,z}(J_v\mid x)$ for every regime $z$, so
\[
  \log\frac{\ell_t(z^{\star})}{\ell_t(z)}
  =\sum_{v\in\calV_k}\ind{v\in\calI(A)}\,
   \log\frac{\pi_{v,z^{\star}}(J_v\mid x)}{\pi_{v,z}(J_v\mid x)} .
\]
Take expectations termwise. Identity \eqref{eq:reach} at regime $z^{\star}$ states that the joint probability of reaching $v$ and taking child $j$ there factors as $\rho_{v,z^{\star}}(x)\,\pi_{v,z^{\star}}(j\mid x)$; equivalently, conditioned on the event $\{v\in\calI(A)\}$, whose probability is $\rho_{v,z^{\star}}$, the child $J_v$ has law $\pi_{v,z^{\star}}(\cdot\mid x)$. Hence for any function $g$, $\E[\ind{v\in\calI(A)}\,g(J_v)] =\rho_{v,z^{\star}}\sum_j\pi_{v,z^{\star}}(j\mid x)\,g(j)$. With $g$ the logarithmic ratio the inner sum is $\KL(\pi_{v,z^{\star}}\,\|\,\pi_{v,z})$, which proves \eqref{eq:ident}; every term is nonnegative because it is a divergence weighted by a probability. When $\calV_k$ contains all internal nodes, the same computation applied to $\log[p_{\Theta}(A\mid x,z^{\star})/p_{\Theta}(A\mid x,z)] =\sum_{v\in\calI(A)}\log[\pi_{v,z^{\star}}(J_v\mid x)/\pi_{v,z}(J_v\mid x)]$ gives $\KL(p_{\Theta}(\cdot\mid x,z^{\star})\,\|\,p_{\Theta}(\cdot\mid x,z))$. Finally, \eqref{eq:update} implies $\log[b_t^{+}(z^{\star})/b_t^{+}(z)] =\log[b_t^{-}(z^{\star})/b_t^{-}(z)]+\log[\ell_t(z^{\star})/\ell_t(z)]$, so \eqref{eq:ident} is the expected increase of the posterior log-odds produced by the measurement update, and unresolved nodes remove exactly their terms.
\end{proof}

\section[Proof of Theorem 1]{Proof of Theorem~\ref{thm:track}}\label{app:track}
\noindent\textbf{Theorem~\ref{thm:track} (Individual-sequence tracking, restated).} \emph{Let $\widehat p_t$ be the predictor \eqref{eq:pred} produced by \eqref{eq:update}--\eqref{eq:time} with fixed parameters, on the replay sequence. For every context sequence $x_{1:T}$ predictable at each step, meaning each $x_t$ is a function of $h_t^{-}$, every action sequence $a_{1:T}$, and every regime path $z_{1:T}\in\calZ^{T}$,}
\begin{equation*}\tag{\ref*{eq:track}}
\begin{aligned}
  \sum_{t=1}^{T}-\log\widehat p_t(a_t)
  \le
  \sum_{t=1}^{T}&-\log p_{\Theta}(a_t\mid x_t,z_t)\\[-1pt]
  &-\log\nu(z_1)
  -\!\sum_{t=2}^{T}\!\log P_{\eta}(z_t\mid z_{t-1},x_t).
\end{aligned}
\end{equation*}
\emph{No assumption is placed on the process generating $a_{1:T}$: the inequality holds pointwise on sequences, so the deployed filter tracks the best regime path in hindsight, and the fallback state of Corollary~\ref{cor:safe} supplies a nonvacuous comparator even when no learned regime fits.}
\begin{proof}
Fix the realized observer data and the scored sequence $a_{1:T}$. Predictability means that $x_t$ and the masks at step $t$ are fixed functions of the realized prefix $h_t^{-}$, which contains $a_{1:t-1}$ in replay; in particular no $x_s$ depends on the hidden path or on actions after step $s$. Define $m_0:=1$ and, for $t\ge1$,
\[
  m_t:=\sum_{z_{1:t}}\nu(z_1)\prod_{s=2}^{t}P_{\eta}(z_s\mid z_{s-1},x_s)
  \prod_{s=1}^{t}p_{\Theta}(a_s\mid x_s,z_s),
\]
the model probability of $a_{1:t}$ with every factor evaluated at the realized inputs. Introduce the unnormalized forward variables $\alpha_1(z):=\nu(z)\,p_{\Theta}(a_1\mid x_1,z)$ and, for $t\ge2$, $\alpha_t(z):=p_{\Theta}(a_t\mid x_t,z)\sum_{z'}P_{\eta}(z\mid z',x_t)\, \alpha_{t-1}(z')$; summing out the hidden path gives $m_t=\sum_z\alpha_t(z)$. We claim that the recursion \eqref{eq:update}--\eqref{eq:time}, run with the replay evidence $\ell_t(z)=p_{\Theta}(a_t\mid x_t,z)$, satisfies $b_t^{+}=\alpha_t/m_t$ and $\widehat p_t(a_t)=m_t/m_{t-1}$ for every $t$. For $t=1$, $b_1^{-}=\nu$ gives $\widehat p_1(a_1)=\sum_z\nu(z)\,p_{\Theta}(a_1\mid x_1,z)=m_1=m_1/m_0$ and $b_1^{+}(z)=\ell_1(z)\,\nu(z)/m_1=\alpha_1(z)/m_1$. Assume $b_{t-1}^{+}=\alpha_{t-1}/m_{t-1}$. Then \eqref{eq:time} gives $b_t^{-}(z)=\sum_{z'}P_{\eta}(z\mid z',x_t)\,\alpha_{t-1}(z')/m_{t-1}$, hence by \eqref{eq:pred}
\[
\begin{aligned}
  \widehat p_t(a_t)
  &=\frac{1}{m_{t-1}}\sum_z p_{\Theta}(a_t\mid x_t,z)
   \sum_{z'}P_{\eta}(z\mid z',x_t)\,\alpha_{t-1}(z')\\
  &=\frac{m_t}{m_{t-1}},
\end{aligned}
\]
and \eqref{eq:update} gives $b_t^{+}(z)=\ell_t(z)\,b_t^{-}(z)\big/\sum_r\ell_t(r)\,b_t^{-}(r) =\alpha_t(z)/m_t$, closing the induction. The product therefore telescopes,
\begin{equation}
\begin{aligned}
  \prod_{t=1}^{T}\widehat p_t(a_t)=m_T
  =\sum_{z'_{1:T}}\nu(z'_1)&\prod_{t\ge2}P_{\eta}(z'_t\mid z'_{t-1},x_t)\\[-1pt]
  &\times\prod_{t=1}^{T}p_{\Theta}(a_t\mid x_t,z'_t),
\end{aligned}
\label{eq:mixture}
\end{equation}
and every summand of \eqref{eq:mixture} is nonnegative, so $m_T$ is at least the single summand indexed by the comparator path $z_{1:T}$. Taking negative logarithms yields \eqref{eq:track}. All quantities are finite whenever $\nu(z_1)>0$: the conservative masks keep $a_t\in\calA_t$, so every action-model factor is positive, and the kernels are strictly positive. No property of the mechanism generating $a_{1:T}$ was used, so the inequality holds pointwise on sequences.
\end{proof}
The argument is the standard evaluation of a Bayesian mixture by the forward algorithm, as in the analysis of tracking a changing best expert~\cite{herbster1998,cesabianchi2006}; it is recorded in full because the input-dependent kernel and the mask-dependent action model enter the bookkeeping. The same telescoping yields the corresponding guarantee for the detection evidence of Section~\ref{sec:model}. Let the channel satisfy $\sum_e\omega(e\mid a)=1$, and let $\widehat q_t:=\sum_z b_t^{-}(z)\,\ell_t(z)$ be the predictive probability of the received evidence. Replacing every action-model factor by the evidence weight $\ell_t$ in the argument above gives, for every comparator path,
\[
\begin{aligned}
  \sum_{t=1}^{T}-\log\widehat q_t
  \le\sum_{t=1}^{T}&-\log\ell_t(z_t)\\[-1pt]
  &-\log\nu(z_1)-\!\sum_{t=2}^{T}\!\log P_{\eta}(z_t\mid z_{t-1},x_t).
\end{aligned}
\]

\section[Proof of Corollary 2]{Proof of Corollary~\ref{cor:safe}}\label{app:cor2}
\noindent The two bounds of Corollary~\ref{cor:safe} are proved in turn.

\medskip\noindent\textbf{Corollary~\ref{cor:safe}, fixed-share bound (restated).} \emph{With uniform $\nu$ and the homogeneous kernel \eqref{eq:sticky}, a comparator path with $S$ switches gives}
\begin{equation*}\tag{\ref*{eq:sticky_bound}}
\begin{aligned}
  \sum_{t=1}^{T}-\log\widehat p_t(a_t)
  \le
  \sum_{t=1}^{T}&-\log p_{\Theta}(a_t\mid x_t,z_t)+\log K\\[-1pt]
  &+(T{-}1{-}S)\log\tfrac1\kappa+S\log\tfrac{K-1}{1-\kappa}.
\end{aligned}
\end{equation*}
%
\emph{The bound is the fixed-share guarantee for tracking a changing best expert~\cite{herbster1998,cesabianchi2006}, with the regime-conditioned choice laws as the experts.}
\begin{proof}
Under uniform $\nu$, $-\log\nu(z_1)=\log K$. Under \eqref{eq:sticky} the comparator's $T-1$ transitions comprise $T-1-S$ self-transitions, each contributing $\log\tfrac1\kappa$, and $S$ switches, each contributing $\log\tfrac{K-1}{1-\kappa}$; substituting into \eqref{eq:track} gives \eqref{eq:sticky_bound}.
The resulting guarantee coincides with the fixed-share bound for tracking a changing best expert~\cite{herbster1998}.
\end{proof}

\medskip\noindent\textbf{Corollary~\ref{cor:safe}, uniform fallback (restated).}
\emph{With $\bot$ an ordinary state of the homogeneous kernel \eqref{eq:sticky} on the $K{+}1$ regimes and uniform $\nu$, the constant comparator $z_t\equiv\bot$ in \eqref{eq:track} gives, for every action sequence,}
\begin{equation*}\tag{\ref*{eq:safe}}
  \sum_{t=1}^{T}-\log\widehat p_t(a_t)\le\sum_{t=1}^{T}\log|\calA_t|
  +\log(K{+}1)+(T{-}1)\log\tfrac1\kappa,
\end{equation*}
\emph{the $S{=}0$ case of \eqref{eq:sticky_bound} at comparator $\bot$, so the mean loss exceeds uniform coding by at most $\log\tfrac1\kappa+\tfrac{\log(K+1)}{T}$. Relative to the model without $\bot$, the right side of \eqref{eq:sticky_bound} for a comparator path through the learned regimes with $S$ switches increases by exactly $\log\tfrac{K+1}{K}+S\log\tfrac{K}{K-1}$ over the episode.}
\begin{proof}
First, the background regime is well defined on the pruned tree. For an internal node $v$, the feasible leaves below $v$ partition among its feasible children, so $\sum_{j\in\calC_t(v)}N_t(j)=N_t(v)$ and the weights $\pi_{v,\bot}(j\mid x)=N_t(j)/N_t(v)$ are positive and sum to one; along the path of any $a\in\calA_t$ the ratios telescope, giving $p_{\Theta}(a\mid x,\bot)=1/N_t(\mathrm{root})=1/|\calA_t|$, while $\bot$ retains the product form \eqref{eq:leaf} and hence Lemma~\ref{lem:prefix}. Apply \eqref{eq:track} on the homogeneous kernel over $K+1$ regimes with uniform $\nu$ and the constant comparator $z_t\equiv\bot$: the initial term is $\log(K{+}1)$, all $T-1$ transitions are self-transitions contributing $\log\tfrac1\kappa$ each, and the emission terms are $\sum_t\log|\calA_t|$; this is \eqref{eq:safe} and is the $S{=}0$ case of \eqref{eq:sticky_bound} on $K{+}1$ states. Dividing by $T$ and using $(T-1)/T\le1$ bounds the mean loss by the mean of $\log|\calA_t|$ plus $\log\tfrac1\kappa+\tfrac{\log(K+1)}{T}$. For the last claim, fix a comparator path through the $K$ learned regimes with $S$ switches and compare the right side of \eqref{eq:sticky_bound} on $K{+}1$ states with the same quantity on $K$ states at the same $\kappa$: the emission terms and the $T{-}1{-}S$ self-transition terms agree, the initial term grows by $\log\tfrac{K+1}{K}$, and each switch term grows by $\log\tfrac{K}{1-\kappa}-\log\tfrac{K-1}{1-\kappa}=\log\tfrac{K}{K-1}$, giving the stated increase.
\end{proof}

\section{Adaptation Delay After a Switch}\label{app:delay}
\noindent The following remark is referenced at the end of Section~\ref{sec:theory}, where only its conclusion is quoted.

\begin{remark}[Adaptation delay]\label{rem:delay}
Suppose the regime switches to $z^{\star}$ and subsequent actions are drawn from $p_{\Theta}(\cdot\mid x_t,z^{\star})$ and fully observed. Kernel \eqref{eq:sticky} keeps every $b_t^{-}(z)\ge(1-\kappa)/(K-1)$, so the initial log-odds deficit of $z^{\star}$ against a fixed regime $z$ is at most $\log\tfrac{K-1}{1-\kappa}$. Before $z^{\star}$ overtakes $z$ the homogeneous time update cannot increase this deficit, while each fully observed action contributes the expected evidence in \eqref{eq:ident}. If that rate remains at least $r>0$ and each single-step log-likelihood ratio has magnitude at most $B$, the expected pairwise crossing time is at most $\bigl(\log\tfrac{K-1}{1-\kappa}+B\bigr)/r$.
\end{remark}

\begin{proof}
Fix $z\neq z^{\star}$ and let $D_t=\log[b_t^{-}(z)/b_t^{-}(z^{\star})]$ denote the log-odds deficit of $z^{\star}$, with $t=0$ the first prediction after the switch. The time update under \eqref{eq:sticky} leaves every regime mass at least $(1-\kappa)/(K-1)$, so $D_0\le\log\frac{K-1}{1-\kappa}$.

One step consists of the measurement \eqref{eq:update} followed by the time update \eqref{eq:time}. The measurement adds $-\Lambda_t$ to the log-odds, where $\Lambda_t=\log[\ell_t(z^{\star})/\ell_t(z)]$. Under \eqref{eq:sticky} the time update acts on each belief coordinate as $b\mapsto\alpha b+\beta$ with $\alpha=\kappa-\frac{1-\kappa}{K-1}\ge0$ and $\beta=\frac{1-\kappa}{K-1}>0$. For $b_1,b_2>0$ the ratio $(\alpha b_1+\beta)/(\alpha b_2+\beta)$ is a weighted mediant of $b_1/b_2$ and $1$ and lies between them, so the time update moves the log-odds toward zero and preserves their sign.

Let $\tau=\min\{t\ge1:D_t\le0\}$. While $\tau>t{+}1$ the post-measurement odds are still positive, the time update shrinks them, and $D_{t+1}\le D_t-\Lambda_t$; by induction $D_t\le D_0-\sum_{s<t}\Lambda_s$ for $t<\tau$. Hence $\tau\le\tau'$, where $\tau'$ is the first time the partial sums of the $\Lambda_s$ exceed $D_0$.

The action is drawn from $p_{\Theta}(\cdot\mid x_t,z^{\star})$, so Corollary~\ref{cor:ident} gives $\E[\Lambda_t\mid\calF_t^{-}]=r(x_t)$. Write $r:=\inf_t r(x_t)>0$ for the assumed rate floor and let $B$ bound $|\Lambda_t|$. The sums $\sum_{s<t}(\Lambda_s-r(x_s))$ form a martingale with increments in $[-2B,2B]$, and Azuma's inequality gives $\Pr(\tau'>t)\le\exp\{-(rt-D_0)_+^{2}/(8tB^{2})\}$, so $\E[\tau']<\infty$. The stopped sum exceeds $D_0$ by at most one increment, so optional stopping gives $r\,\E[\tau']\le\E\bigl[\sum_{s<\tau'}\Lambda_s\bigr]\le D_0+B$. Therefore $\E[\tau]\le(D_0+B)/r\le\bigl(\log\tfrac{K-1}{1-\kappa}+B\bigr)/r$.
\end{proof}

\medskip
\noindent Both constants are finite in the setting of Section~\ref{sec:exp}: the feasible tree has bounded depth and the utilities \eqref{eq:score} are bounded on the finite descriptor set, so the per-node log-ratios, and hence $\Lambda_t$, are bounded, giving a finite $B$; and whenever the post-switch regime $z^{\star}$ is separated from $z$ at some reachable node, the corresponding term of \eqref{eq:ident} is strictly positive, giving $r>0$.

\section{Cost of Committing to One Regime}\label{app:marg}
\noindent Appendices~\ref{app:marg} and~\ref{app:kl} record two results that quantify, respectively, the cost of collapsing the posterior average to a single regime and the exact per-step divergence of the joint model. Neither is used in the body; both are stated and proved here in full.

\begin{proposition}[Cost of committing to one regime]\label{prop:commit}
Fix $x$ and a belief $b$ with $b(\widehat z)<1$; write $p_z=p_{\Theta}(\cdot\mid x,z)$, $\bar p=\sum_z b(z)p_z$, $\widehat z\in\argmax_z b(z)$, let $\bar p_{-\widehat z}$ be the remainder renormalized, and let $G_{\mathrm{cmt}}=\KL(\bar p\,\|\,p_{\widehat z})$ be the excess log-loss of committing to $\widehat z$. Then $\bar p$ uniquely minimises $\E_{\bar p}[-\log q(A)]$, the excess of any $q$ is $\KL(\bar p\,\|\,q)$, and
\begin{equation}\label{eq:twosided}
  2\bigl(1-b(\widehat z)\bigr)^{2}
  \TV^{2}\!\bigl(p_{\widehat z},\bar p_{-\widehat z}\bigr)
  \le G_{\mathrm{cmt}}\le
  \sum_z b(z)\,\KL\bigl(p_z\,\|\,p_{\widehat z}\bigr).
\end{equation}
\end{proposition}
\begin{proof}
Every $p_z$ is positive on $\calA_t$, hence so is $\bar p$, and $\bar p_{-\widehat z} =(\bar p-b(\widehat z)\,p_{\widehat z})/(1-b(\widehat z))$ is a probability distribution because $b(\widehat z)<1$ and $\bar p\ge b(\widehat z)\,p_{\widehat z}$ pointwise. For any distribution $q$ on $\calA_t$, $\E_{\bar p}[-\log q(A)]=H(\bar p)+\KL(\bar p\,\|\,q)$, which is minimized exactly when $\KL(\bar p\,\|\,q)=0$, that is at $q=\bar p$, and the excess of any $q$ is $\KL(\bar p\,\|\,q)$; taking $q=p_{\widehat z}$ identifies $G_{\mathrm{cmt}}$, finite by positivity of $p_{\widehat z}$. For the lower bound, $\bar p-p_{\widehat z} =(1-b(\widehat z))(\bar p_{-\widehat z}-p_{\widehat z})$ pointwise, so $\TV(\bar p,p_{\widehat z}) =(1-b(\widehat z))\,\TV(p_{\widehat z},\bar p_{-\widehat z})$, and Pinsker's inequality gives $G_{\mathrm{cmt}}\ge2\,\TV^{2}(\bar p,p_{\widehat z}) =2(1-b(\widehat z))^{2}\,\TV^{2}(p_{\widehat z},\bar p_{-\widehat z})$. For the upper bound, apply the log-sum inequality at each $a$ with $a_z=b(z)\,p_z(a)$ and $b_z=b(z)\,p_{\widehat z}(a)$:
\[
  \bar p(a)\log\frac{\bar p(a)}{p_{\widehat z}(a)}
  \le\sum_z b(z)\,p_z(a)\log\frac{p_z(a)}{p_{\widehat z}(a)} ,
\]
and summing over $a$ gives $G_{\mathrm{cmt}}\le\sum_z b(z)\,\KL(p_z\,\|\,p_{\widehat z})$.
\end{proof}

\section{A Per-Step Divergence Identity}\label{app:kl}
\begin{proposition}[Per-step divergence]\label{prop:kl}
Fix $x$ and a common feasible tree. For joint laws $P(z,a)=w(z)\prod_{(v,j_v(a))\in\Path(a)}\pi_{v,z}(j_v(a)\mid x)$ with belief vectors $w^{\star},\widehat w$ and reach probabilities $\rho^{\star}_{v,z}=P^{\star}(v\in\calI(A)\mid z)$,
\begin{equation}\label{eq:kl}
\begin{aligned}
  \KL\bigl(P^{\star}_{ZA}\,\|\,\widehat P_{ZA}\bigr)
  =\;&\KL\bigl(w^{\star}\|\widehat w\bigr)\\
  &+\sum_z w^{\star}(z)\sum_{v\in\calI}\rho^{\star}_{v,z}\,
   \KL\bigl(\pi^{\star}_{v,z}\,\|\,\widehat\pi_{v,z}\bigr),
\end{aligned}
\end{equation}
and $\KL(P^{\star}_A\|\widehat P_A)\le\KL(P^{\star}_{ZA}\|\widehat P_{ZA})$.
\end{proposition}
\begin{proof}
Insert $P(z,a)=w(z)\,P(a\mid z)$ and split the logarithm:
\begin{align*}
  \KL\bigl(P^{\star}_{ZA}\,\|\,\widehat P_{ZA}\bigr)
  &=\sum_z w^{\star}(z)\log\frac{w^{\star}(z)}{\widehat w(z)}\\
  &\quad+\sum_z w^{\star}(z)\,
    \KL\bigl(P^{\star}_{A\mid z}\,\|\,\widehat P_{A\mid z}\bigr),
\end{align*}
with all terms finite because $\widehat w$ and the local laws are positive on the common feasible tree. Fix $z$ and let $A\sim P^{\star}_{A\mid z}$. As in the proof of Corollary~\ref{cor:ident},
\[
  \log\frac{P^{\star}(A\mid z)}{\widehat P(A\mid z)}
  =\sum_{v\in\calI}\ind{v\in\calI(A)}\,
  \log\frac{\pi^{\star}_{v,z}(j_v(A)\mid x)}
           {\widehat\pi_{v,z}(j_v(A)\mid x)},
\]
and taking expectations termwise, using \eqref{eq:reach} at the law $\pi^{\star}_{v,z}$ exactly as in the proof of Corollary~\ref{cor:ident} to factor each term into the reach probability $\rho^{\star}_{v,z}$ times a local divergence, gives $\KL(P^{\star}_{A\mid z}\,\|\,\widehat P_{A\mid z}) =\sum_{v\in\calI}\rho^{\star}_{v,z}\, \KL(\pi^{\star}_{v,z}\,\|\,\widehat\pi_{v,z})$, which is \eqref{eq:kl}. The final inequality is the chain rule in the other direction, $\KL(P^{\star}_{ZA}\,\|\,\widehat P_{ZA}) =\KL(P^{\star}_{A}\,\|\,\widehat P_{A}) +\E_{P^{\star}_{A}}\bigl[\KL\bigl(P^{\star}_{Z\mid A}\,\|\, \widehat P_{Z\mid A}\bigr)\bigr] \ge\KL(P^{\star}_{A}\,\|\,\widehat P_{A})$.
\end{proof}

\section{Details Referenced from the Body}\label{app:verify}
\noindent This appendix supplies the derivations, identity checks, and experimental detail that the body asserts or references without developing. The items are grouped by the part of the paper they support.

\subsection{Model, inference, and estimation}

\emph{The persistence mapping (Section~\ref{sec:problem}).} With $\eta\equiv0$ the numerator of \eqref{eq:transition} equals $e^{c}$ at $z'=z$ and $1$ otherwise, and the denominator equals $e^{c}+K-1$, so the self-transition probability is $\kappa=e^{c}/(e^{c}+K-1)$ and every off-diagonal entry is $1/(e^{c}+K-1)=(1-\kappa)/(K-1)$; moreover $c\ge0$ holds exactly when $\kappa\ge1/K$, and $\kappa\uparrow1$ as $c\to\infty$. Under \eqref{eq:sticky} the holding time of a regime is geometric with parameter $1-\kappa$, hence the mean dwell $1/(1-\kappa)$.

\emph{The static restriction at $c=0$ (Section~\ref{sec:problem}).} At $c=0$ all rows of \eqref{eq:transition} coincide with $\mathrm{softmax}_{z'}(\eta_{z'}^{\top}\bar x_t)$, so for $t\ge2$ the time update \eqref{eq:time} returns that gate for every $b_{t-1}^{+}$; after the common initialization $\nu$ at $t=1$ the filter carries no information across steps and the predictor reduces to the static contextual mixture. The persistent and static models are therefore nested and are compared by held-out predictive NLL in Section~\ref{sec:exp}.

\emph{Identification of the transition parameters (Section~\ref{sec:problem}).} The destination coefficients of \eqref{eq:transition} are determined only up to a common shift, so we set $\eta_{z_{\mathrm{ref}}}=0$ for one reference regime. Using destination-specific context effects together with a single shared persistence term $c$ keeps the transition model at $K$ context vectors rather than $K^{2}$, which matters because switch data are sparse.

\emph{The mask decomposition (Section~\ref{sec:problem}).} Let $S_t$ be the true support of $A_t$ and $\calA_t\supseteq S_t$ the conservative feasible set. For $a_t\in S_t$ the law renormalized to $S_t$ satisfies $\widehat p_t(a_t)=\widehat p_t^{S}(a_t)\,\widehat p_t(S_t)$, so the log loss splits into the loss on the true support and the log-probability the model assigns to it; a mask that excluded $a_t$ would force $\widehat p_t(a_t)=0$ by \eqref{eq:leaf} and infinite loss, which is why the masks are conservative.

\emph{Identifiability through the residual (Section~\ref{sec:model}).} If $\Delta_{v,z}\equiv0$ then $p_{\Theta}(\cdot\mid x,z)$ does not depend on $z$, every evidence weight $\ell_t$ is constant in $z$, \eqref{eq:update} returns $b_t^{+}=b_t^{-}$, and the posterior equals the prior propagated by \eqref{eq:time}: no action sequence moves it.

\emph{Convexity, separability, and uniqueness of the estimator (Section~\ref{sec:model}).} Each summand of \eqref{eq:supervised} has the form $\log\sum_k\exp u_k-u_j$ with utilities affine in the parameters, the composition of the convex log-sum-exp function with an affine map, hence convex; the objective is a sum of such terms and quadratics, is jointly convex, and separates because the parameters of node $v$ enter only the terms of node $v$ and $(c,\eta)$ only the transition terms. The ridge penalties make each block strictly convex, so its minimiser is unique; the constraints $\sum_z\Delta_{v,z}=0$, the reference child, and the centered descriptors remove the directions the likelihood cannot see, so that $\theta_v^{0}$ is the across-regime mean preference. In Baum--Welch the maximisation step solves the responsibility-weighted versions of the same convex blocks, and the usual EM argument gives monotone ascent of the penalized marginal likelihood to a stationary point~\cite{cappe2005}.

\emph{Sufficiency of the belief (Section~\ref{sec:model}).} Under the model, the conditional law of $(Z_{t+1},A_{t+1})$ given $\calF_t^{+}$ depends on the history only through $b_t^{+}$ and the forthcoming inputs, because \eqref{eq:time} maps $(b_t^{+},x_{t+1})$ to $b_{t+1}^{-}$ and the emission depends on $(x_{t+1},Z_{t+1})$ alone; iterating, $b_t^{+}$ is a sufficient statistic of the history for the regime given the context process.

%

\emph{The per-step cost (Section~\ref{sec:model}).} One filtering step costs $O(K^{2})$ for the time update \eqref{eq:time} under the general kernel \eqref{eq:transition} and $O\!\bigl(K\sum_{v}(d\,r_v+|\calC_t(v)|\,r_v)\bigr)$ for the tree traversals. Reshaping $\theta_v^{0}+\Delta_{v,z}$ as a $(d{+}1)\times r_v$ matrix $\Theta_{v,z}$ gives $u_{v,z,j}(x)=\bar x^{\top}\Theta_{v,z}e_{v,j}$; computing $g:=\Theta_{v,z}^{\top}\bar x$ once costs $O(d\,r_v)$ and each child score $g^{\top}e_{v,j}$ costs $O(r_v)$, giving the stated $O(d\,r_v+|\calC_t(v)|\,r_v)$ per node and regime, while under \eqref{eq:sticky} the time update is the affine map $b^{-}(z')=\bigl(\kappa-\tfrac{1-\kappa}{K-1}\bigr)b^{+}(z') +\tfrac{1-\kappa}{K-1}$, computed in $O(K)$.

\subsection{Analysis and guarantees}

\emph{Second-order form of the evidence rate (Corollary~\ref{cor:ident}, Section~\ref{sec:theory}).}
Fix a node $v$ and write $u_j=u_{v,z^{\star},j}(x)$ and $u_j-\delta_j=u_{v,z,j}(x)$, so that $\delta_j=\langle\Delta_{v,z^{\star}}-\Delta_{v,z},\phi_v(x,e_{v,j})\rangle$ by \eqref{eq:score}. With $p=\pi_{v,z^{\star}}(\cdot\mid x)=\mathrm{softmax}(u)$ and $q=\pi_{v,z}(\cdot\mid x)=\mathrm{softmax}(u-\delta)$, the log-partition expansion $\log\sum_j e^{u_j-\delta_j}=\log\sum_j e^{u_j}-\E_p[\delta]+\tfrac12\Var_p(\delta)+O(\|\delta\|^3)$ gives $\KL(p\,\|\,q)=\E_p[\delta]+\log\textstyle\sum_j e^{u_j-\delta_j}-\log\sum_j e^{u_j}=\tfrac12\Var_p(\delta)+O(\|\delta\|^3)$. Substituting $\delta_j=\langle\Delta_{v,z^{\star}}-\Delta_{v,z},\phi_v(x,e_{v,j})\rangle$ and weighting by the reach $\rho_{v,z^{\star}}$ turns each term of \eqref{eq:ident} into $\tfrac12\rho_{v,z^{\star}}(\Delta_{v,z^{\star}}-\Delta_{v,z})^{\top}\Cov_{\pi_{v,z^{\star}}}\!\bigl(\phi_v(x,e_{v,\cdot})\bigr)(\Delta_{v,z^{\star}}-\Delta_{v,z})$, a quadratic form in the residual differences; the ridge on $\Delta$ therefore shrinks the residual differences and, through this quadratic form, lowers the identification rate directly, which is the trade between pooling and identification discussed in Section~\ref{sec:model}.

\emph{The hard-commitment comparison (Section~\ref{sec:theory}).} Since \eqref{eq:pred} averages nonnegative terms, $\widehat p_t(a)\ge b_t^{-}(z)\,p_{\Theta}(a\mid x_t,z)$ for every $z$, whence $-\log\widehat p_t(a)\le-\log p_{\Theta}(a\mid x_t,z)+\log(1/b_t^{-}(z))$: the posterior average never loses more than $\log(1/b_t^{-}(z))$ per step against any single regime it still gives weight to.

\emph{A worked example for Corollary~\ref{cor:safe} (Section~\ref{sec:theory}).} At $T=100$, $K=3$, and $\kappa=0.99$, the fallback bound \eqref{eq:safe} exceeds uniform coding over the feasible sets by at most $\log\tfrac1{0.99}+\tfrac{\log4}{100}=0.01005+0.01386<0.024$ nats per step. At the same $K$ and one switch, carrying $\bot$ raises the learned-regime penalty of \eqref{eq:sticky_bound} by $\log\tfrac{K+1}{K}+\log\tfrac{K}{K-1}=\log\tfrac43+\log\tfrac32=\log2=0.693\ldots$ nats per episode, the $S{=}1$ instance of the premium $\log\tfrac{K+1}{K}+S\log\tfrac{K}{K-1}$. A single $\kappa$ therefore buys a cheap worst-case guarantee at a small cost to the switching comparator.

\emph{Optimality of the posterior weighting (after Proposition~\ref{prop:kl}).} Under the model, $\Pr(A_t=a\mid\calF_t^{-}) =\sum_z\Pr(Z_t=z\mid\calF_t^{-})\,p_{\Theta}(a\mid x_t,z)=\widehat p_t(a)$. For any $\calF_t^{-}$-measurable weighting $w$, the mixture $q_w:=\sum_z w(z)\,p_z$ is $\calF_t^{-}$-measurable, so $\E[-\log q_w(A_t)\mid\calF_t^{-}] =H(\widehat p_t)+\KL(\widehat p_t\,\|\,q_w)$; subtracting the same identity at $q=\widehat p_t$ and taking expectations gives $\E[-\log q_w(A_t)]-\E[-\log\widehat p_t(A_t)] =\E[\KL(\widehat p_t\,\|\,q_w)]\ge0$.

\emph{The Dobrushin coefficient (Section~\ref{sec:theory}).} Two distinct rows of \eqref{eq:sticky} differ only in the two coordinates where one places $\kappa$ and the other $(1-\kappa)/(K-1)$, so their total variation is $\kappa-(1-\kappa)/(K-1)=(K\kappa-1)/(K-1)=:\delta(P)$. For beliefs $\mu,\nu$ with $\varepsilon:=\TV(\mu,\nu)$ write $\mu-\nu=\varepsilon(p-q)$ with $p,q$ the normalized positive and negative parts; for any coupling $\gamma$ of $(p,q)$, $P^{\top}\mu-P^{\top}\nu =\varepsilon\sum_{z,z'}\gamma(z,z')\bigl(P(\cdot\mid z)-P(\cdot\mid z')\bigr)$, so $\TV(P^{\top}\mu,P^{\top}\nu)\le\delta(P)\,\TV(\mu,\nu)$; geometric forgetting under bounded action likelihood ratios follows as in~\cite{cappe2005}.

\subsection{Experimental detail}

\emph{Model sizes (Table~\ref{tab:main}, Section~\ref{sec:exp}).} Parameter counts are $3.1$k for the action bigram, $7.3$k for the masked GRU, $7.4$k for the hierarchical choice model and $8.7$k with the previous action in the context, $10.6$k for the flat softmax over the $56$ actions, and $1.5$k for the policy-class chain and $2.8$k with the previous action. The hierarchical model therefore uses about $30\%$ fewer parameters than the flat softmax on the same inputs and is within $2\%$ of the GRU it is compared with.

\emph{Training-set size (Section~\ref{sec:exp}).} Halving the training set to $60$ episodes, from $120$, the full-context choice model reaches $1.007$ nats and $69.5\%$ top-$1$ against $0.996$ nats and $68.3\%$ for the GRU, so the ordering of the two at $120$ episodes reverses in likelihood and holds in accuracy. The hierarchical factorization is at or below the flat softmax up to $60$ episodes ($1.028$ against $1.068$ nats) and behind it from $90$ episodes on, which is the expected pattern for a structural restriction that pays off when data are scarce.

\emph{Unseen attacker (Section~\ref{sec:exp}).} On episodes generated by a red controller absent from training, every learned predictor is overconfident, with ECE above $0.35$. The action bigram degrades least, to $3.43$ nats per step. The GRU reaches $4.48$ nats and the choice models with the filtered posterior reach $5.39$ (flat) and $7.19$ (hierarchical, full context), all worse than the $4.03$ nats of uniform coding over the $56$ actions and worse than simply repeating the last action ($3.95$). This is the failure mode the fallback state $\bot$ of Section~\ref{sec:model} is built for: in the joint filter, Corollary~\ref{cor:safe} caps the loss at uniform coding over the feasible sets plus the initial and persistence penalties, so no such sequence can be catastrophic. The two-stage estimator evaluated here does not carry $\bot$, so this experiment establishes the failure mode and leaves the guarantee to be tested with the joint filter.